\documentclass[runningheads,orivec,a4paper]{llncs}

\usepackage{amssymb}
\usepackage{graphicx}
\usepackage{mathtools}
\usepackage{textcomp}
\usepackage{tikz-cd}

\begin{document}

\mainmatter  

\title{Scalable multimodal convolutional networks\\
	for brain tumour segmentation} 

\titlerunning{Scalable multimodal convolutional networks}

\author{Lucas Fidon\inst{1}%
\and Wenqi Li\inst{1}\and Luis C. Garcia-Peraza-Herrera\inst{1}\and\\ 
Jinendra Ekanayake\inst{2,3}\and Neil Kitchen\inst{2}\and\\Sebastien Ourselin\inst{1,3}\and Tom Vercauteren\inst{1,3}}


\authorrunning{Lucas Fidon et al.}

\institute{TIG, CMIC, University College London, London, UK\\
\and
NHNN, University College London Hospitals, London UK\\
\and
Wellcome/EPSRC Centre for Surgical and Interventional Science, \\UCL, London, UK}

\maketitle

\begin{abstract}
Brain tumour segmentation plays a key role in computer-assisted surgery. Deep neural networks have increased the accuracy of automatic segmentation significantly, however these models tend to generalise poorly to different imaging modalities than those for which they have been designed, thereby limiting their applications. For example, a network architecture initially designed for brain parcellation of monomodal T1 MRI can not be easily translated into an efficient tumour segmentation network that jointly utilises T1, T1c, Flair and T2 MRI. To tackle this, we propose a novel scalable multimodal deep learning architecture using new nested structures that explicitly leverage deep features within or across modalities. This aims at making the early layers of the architecture structured and sparse so that the final architecture becomes scalable to the number of modalities. We evaluate the scalable architecture for brain tumour segmentation and give evidence of its regularisation effect compared to the conventional concatenation approach.
\end{abstract}

\section{Introduction}
Gliomas make up 80$\%$ of all malignant brain tumours. Tumour-related tissue changes can be captured by various MR modalities, including T1, T1-contrast, T2, and Fluid Attenuation Inversion Recovery (FLAIR). Automatic segmentation of gliomas from MR images is an active field of research that promises to speed up diagnosis, surgery planning, and follow-up evaluations.
Deep Convolutional Neural Networks (CNNs) have recently achieved state-of-the-art results on this task~\cite{chen2016voxresnet,Havaei201718,kamnitsas2017efficient,pereira2016brain}.
Their success is partly attributed to their ability of automatically learning
hierarchical visual features as opposed to conventional hand-crafted
features extraction.
Most of the existing multimodal network architectures handle imaging
modalities by concatenating the intensities as an input.  The multimodal
information is implicitly fused by training the network discriminatively.
Experiments show that relying on multiple MR modalities consistently is key
to achieving highly accurate segmentations~\cite{havaei2016hemis,menze2015multimodal}. 
However, using classical modality concatenation to turn a given monomodal architecture into a multimodal CNN does not scale well because it either requires to dramatically augment the number of hidden channels and network parameters, or imposes a bottleneck on at least one of the network layers.
This lack of scalability requires the design of dedicated multimodal architectures and makes it
difficult and time-consuming to adapt state-of-the-art network architectures.

Recently, Havaei et al.~\cite{havaei2016hemis} proposed an hetero-modal network architecture (HeMIS)
that learns to embed the different modalities into a common latent space. 
Their work suggests that it is possible to impose more structure on the network. HeMIS separates the CNN into a backend that encodes modality-specific features up to the common latent space, and a frontend that uses high-level modality-agnostic feature abstractions.
HeMIS is able to deal with missing modalities and shows
promising segmentation results. However, the authors do not study the
adaption of existing networks to additional imaging modalities and do not demonstrate an optimal
fusion of information across modalities.

We propose a scalable network framework (ScaleNets) that enables efficient refinement of an existing architecture to adapt it to an arbitrary number of MR modalities instead of building a new architecture from scratch. ScaleNets are CNNs split into a backend and frontend with across-modality information flowing through the backend thereby alleviating the need for a one-shot latent space merging.
The proposed scalable backend takes advantage of a factorisation of the feature space into imaging modalities ($M$-space) and modality-conditioned features ($F$-space).
By explicitly using this factorisation, we impose sparsity on the network structure with demonstrated improved generalisation.


We evaluate our framework by starting from a high-resolution network initially designed for brain parcellation from T1 MRI \cite{highresnet} and readily adapting it to brain tumour segmentation from T1, T1c, Flair and T2 MRI.
Finally, we explore the design of the modality-dependent backend by comparing several important factors, including the number of modality-dependent layers, the merging function, and convolutional kernel sizes.
Our experiments show that the proposed networks are more efficient and scalable than the conventional CNNs and achieve competitive segmentation results on the BraTS 2013 challenge dataset.

\section{Structural transformations across features/modalities}
Concatenating multimodal images as input is the simplest and most common approach in CNN-based segmentation \cite{Havaei201718,kamnitsas2017efficient}.
We emphasise that the complete feature space $FM$ can be factorised into a M-feature space $M$ derived from imaging modalities, and a F-feature space $F$ derived from scan intensity. However the concatenation strategy doesn't take advantage of it.

\begin{figure}[t!]
	\centering
	\includegraphics[width=.85\linewidth]{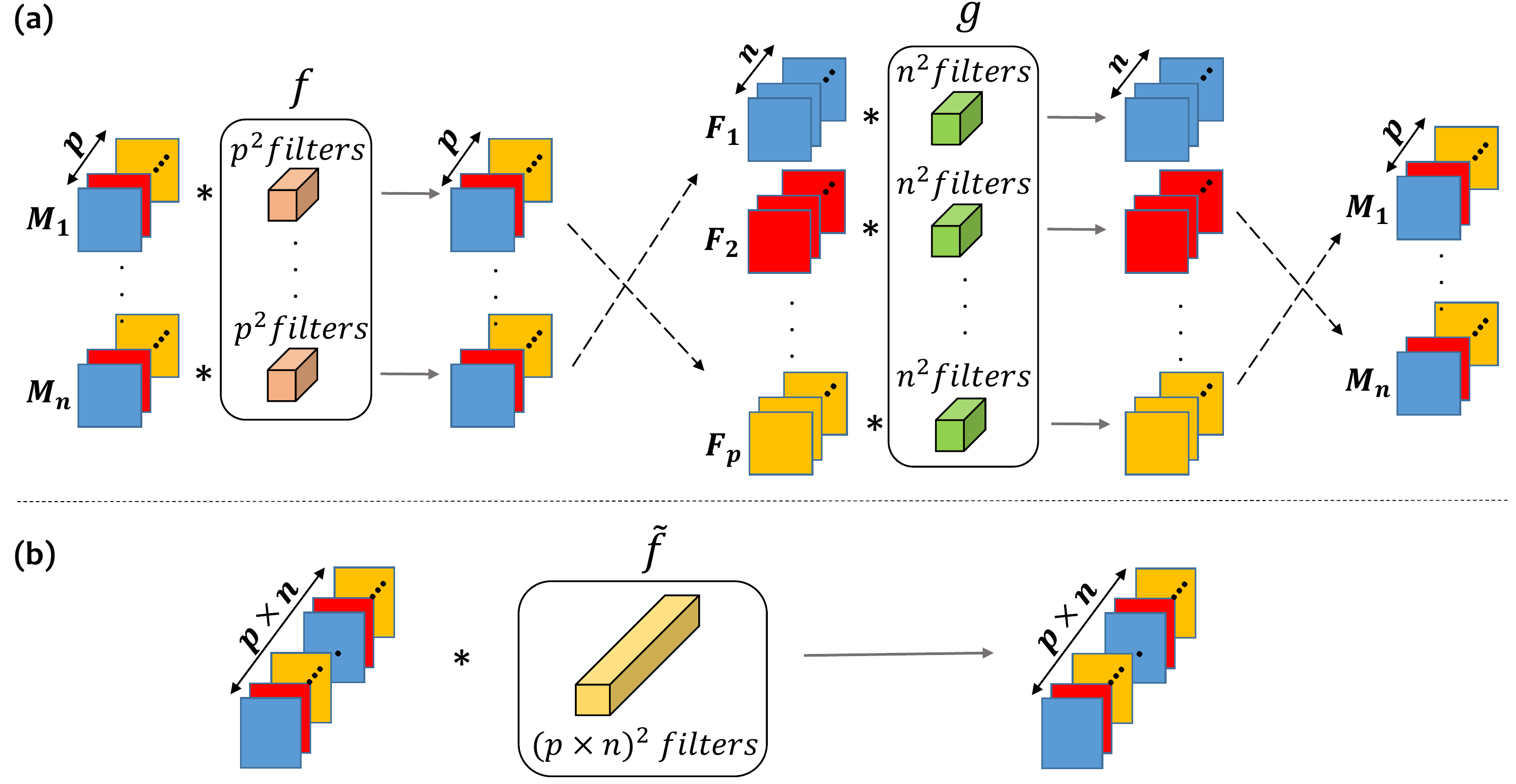}
	\caption{(a) The proposed scalable multimodal layer. (b) A classic CNN layer with multimodal images concatenated as input. 
	Volumes are represented as slices, the colours correspond to the F-features ($F_{1}, ..., F_{p}$) and ($M_1, ..., M_n$) correspond to the M-features. In (a) transformations across F-features $f$ and across M-features $g$ are explicitly separated (as illustrated by the rooted structure) while in (b) there are implicitly both applied in $\hat{f}$.	
    The ratio of the number of parameters in (a) compared to (b) is $\frac{p + n}{p\times n}$.
    }\label{fig:layers}
\end{figure}

We propose to impose structural constraints that make this factorisation explicit.
Let $V \subset \mathbb{R}^{3}$ be a discrete volume domain, and $F$ (resp. $M$) be a finite F-features (resp. M-features) domain, the set of feature maps associated to ($V$, $F$, $M$) is defined as: $\mathcal{G}(V\times F \times M)= \{x:V\times F \times M \rightarrow \mathbb{R}\}$.
This factorisation allows us to introduce new scalable layers that perform the transformation $\tilde{f}$ of the joint $FM$ feature space in two steps (1). $f$ (resp. $g$) typically uses convolutions across $F$-features (resp. across $M$-features).
The proposed layer architecture, illustrated in Fig.~\ref{fig:layers}, offers several advantages compared to classic ones:
(1) cross $F$-feature layers remain to some extent independent of the number of modalities (2) cross $M$-feature layers allow the different modality branches to share complementary information (3) the total number of parameters is reduced. The HeMIS architecture \cite{havaei2016hemis}, where one branch per modality is maintained until averaging merges the branches, is a special case of our framework where the cross M-features transformations $g$ are identity mappings.
\vspace*{-1.5\baselineskip}
\begin{center}
	\begin{equation}
	\begin{tikzcd}
		\mathcal{G}(V\times F \times M) \arrow[rr, dashrightarrow, "\tilde{f}"] \arrow[rd, "f"] &   & \mathcal{G}(V\times F' \times M') \\
		& \mathcal{G}(V\times F' \times M) \arrow[ur, "g"] & 
	\end{tikzcd}
	\end{equation}
\end{center}

Another important component of the proposed framework is the merging layer.
It aims at recombining the F-features space and the M-features space together either by concatenating them or by applying a downsampling/pooling (averaging, maxout) on the M-features space to reduce its dimension to one:
\begin{center}
	   \begin{tikzcd}
		\mathcal{G}(V\times F \times M) \arrow[r, "concat"] & \mathcal{G}(V\times FM), \quad
		\mathcal{G}(V\times F \times M) \arrow[r, "pooling"] & \mathcal{G}(V\times F \times \{1\})
	   \end{tikzcd}
\end{center}
 
%

\noindent As opposed to concatenation, relying on averaging or maxout for the merging layer at the interface between a backend and frontend makes the frontend structurally independent of the number of modalities and more generally of the entire backend. The proposed ScaleNets rely on such merging strategies to offer scalability in the network design.


\section{ScaleNets implementation}
The modularity of the proposed feature factorisation raises different questions:
1) Is the representative power of scalable $F$/$M$-structured multimodal CNN the same as classic ones?
2) What are the important parameters for the tradeoff between accuracy and complexity?
3) How can this modularity help readily transform existing architectures into scalable multimodal ones?

\begin{figure}[b!]
	\centering
	\includegraphics[width=0.9\linewidth]{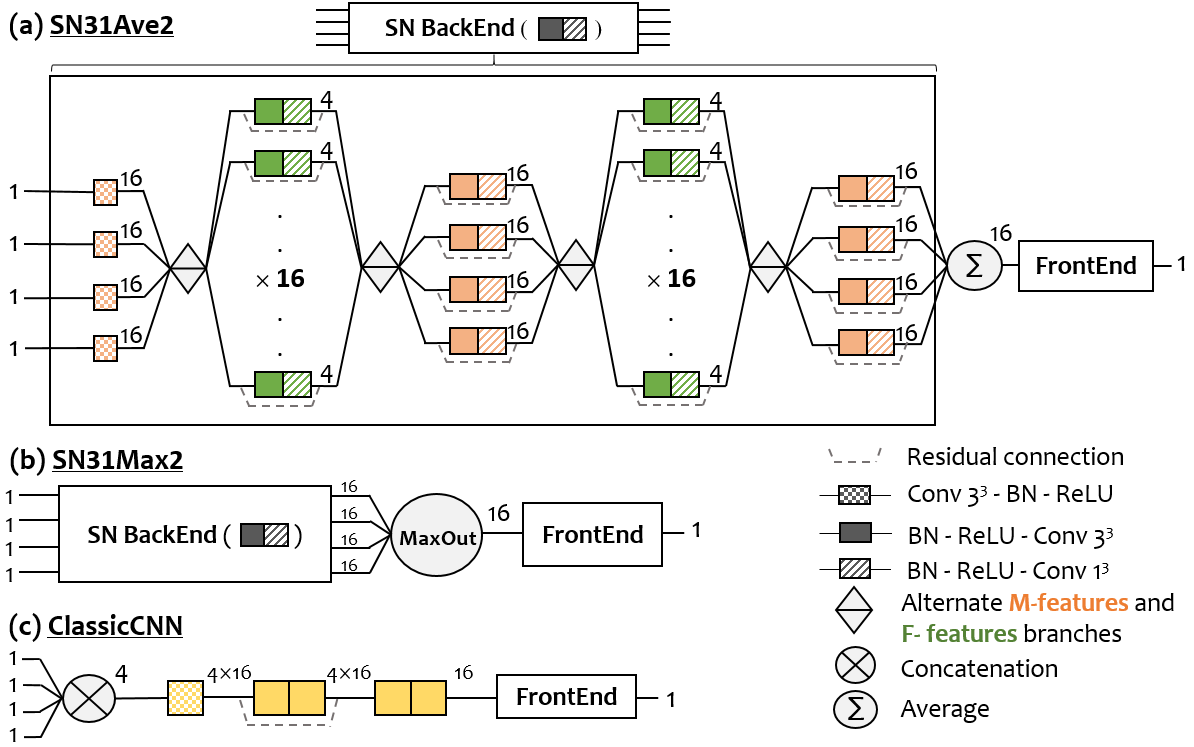}
	\caption{Scalable and Classic CNN architectures. Numbers in bold are for the number of branches and the other correspond to the number of features.}
	\label{fig:expt1}
\end{figure}

To demonstrate that our scalable framework can provide, to a deep network, the flexibility of efficiently being reused for different sets of image modalities, we adapt a model originally built for brain parcellation from T1 MRI~\cite{highresnet}.
As illustrated in Fig.~\ref{fig:expt1}, the proposed ScaleNets splits the network into two parts: (i) a backend and (ii) a frontend.
In following experiments, we explore different backend architectures allowing to scale the monomodal network into a multimodal network. We also add a merging operation that allows plugging any backend into the frontend and makes the frontend independent from the number of modalities used. As a result, the frontend will be the same for all our architectures.

To readily adapt the backend from the monomodal network architecture~\cite{highresnet} we duplicate the layers to get the across $F$-features transformations (one branch per $M$-features) and add an across $M$-features transformation after each of them (one branch per $F$-features) as shown in Fig.~\ref{fig:expt1}.
In the frontend, only the number of outputs of the last layer is changed to match the number of classes for the new task.
The proposed scalable models (SN31Ave1, SN31Ave2, SN31Ave3, SN33Ave2, SN31Max2) are named consistently.
For example, SN31Ave2 stands for: "ScaleNet with 2 cross $M$-features residual blocks with $3^3$ convolution and $1^3$ convolution before averaging" and corresponds to the model (a) of Fig.~\ref{fig:expt1}.

\subsubsection{Baseline monomodal architecture.}
The baseline architecture used for our experiments is a high-resolution, compact network designed for volumetric image segmentation \cite{highresnet}. It has proved to reach state-of-the-art results for brain parcellation of T1 scans.
This fully convolutional neural network makes an end-to-end mapping from a monomodal image volume to a voxel-level segmentation map mainly with convolutional blocks and residual connections. 
It also takes advantage of dilated convolutions to incorporate image features at multiple scales while maintaining the spatial resolution of the input images.
The maximum receptive field
is 87$\times$87$\times$87 voxels and is, therefore, able to catch multi-scale information in one path.
%
By learning the variation between successive feature maps, the residual connections allow the initialisation of cross M-feature transformations closed to identity mappings. Thus it encourages information sharing across the modalities without changing their nature.



\subsubsection{Brain tumour segmentation.}
We compare the different models on the task of brain tumour segmentation using BraTS'15 training set that is composed of 274 multimodal images (T1, T1c, T2 and Flair).  We divide it into $80\%$ for training, $10\%$ for validation and $10\%$ for testing. Additionally, we evaluate one of our scalable network model on the challenge BraTS'13 dataset, for which an online evaluation platform is available\footnote{\url{https://www.virtualskeleton.ch/BraTS/}}, to compare it to state-of-the-art (all the models were trained on the BraTS'15 though).
%



\subsubsection{Implementation details.}
We maximise the soft Dice score as proposed by~\cite{milletari2016v}.
We train all the networks with Adam Optimization method \cite{kingma2014adam} with a learning rate $lr=0.01$, $\beta_1 = 0.9$ and $\beta_2 = 0.999$. We also used early stopping on the validation set.
Rotation of random small angles in the range $[-10\text{\textdegree}, 10\text{\textdegree}]$ are applied along each axis during training. 
All the scans of BraTS dataset are available after skull stripping, resampling to a 1mm isotropic grid and co-registration of all the modalities to the T1-weighted images for each patient.
%
Additionaly, we applied the histogram-based standardisation method~\cite{histStd}.
The experiences have been performed using NiftyNet\footnote{Our implementation of the ScaleNets and other CNNs used for comparison can be found at \url{http://www.niftynet.io}} and one GPU Nvidia GTX Titan.

\subsubsection{Evaluation of segmentation performance.}
Results are evaluated using the Dice score of different tumour subparts: whole tumour, core tumour and enhanced tumour \cite{menze2015multimodal}. Additionally, we introduce a healthy tissue class to separate it from the background (zeroed out in the BraTS dataset).

\section{Experiments and results}

To demonstrate the usefulness of our framework, we compare two basic ScaleNets and a classic CNN.
Tab.~\ref{fig:result1} highlights the benefits of ScaleNets in terms of number of parameters.
We also explore some combinations of the important factors appearing in the choice of the architecture to try to address some key practical questions. How deep does the cross modalities layers have to be? When should we merge the different branches? Which merging operation should we use?
Wilcoxon signed-rank p-values are reported to highlight significant improvements.



\begin{table}[t!]
	\centering
	\caption{Comparison of ScaleNets and Classic concatenation-based CNNs for model adaptation on the testing set.}
	\begin{tabular}{l|c|ccc}
		\hline
		& & \multicolumn{3}{c}{\bf Mean(Std) Dice Score (\%)}\\
		\hline
		Method & \# Param. &\multicolumn{1}{c}{~~Whole Tumour~~}   & \multicolumn{1}{c}{~~Core Tumour~~}   & \multicolumn{1}{c}{~~Active Tumour~~}  \\ 
		\hline
		SN31Ave1   & \bf0.83M & 87(8)& \textbf{73}(22)& \bf72(26)\\
		SN31Ave2   & 0.85M & 87(7) & 71(19) & 70(28) \\
		SN31Ave3   & 0.88M & \bf88(6) & 69\bf(17) & 71(27) \\
		SN31Max2   & 0.85M & 85(9) & 67\bf(17) & 71(28) \\
		SN33Ave2   & 0.92M & 87(7) & 70(18) & 67(27) \\
		HeMIS-like  & 0.89M & 86(12) & 70(20) & 69(28) \\
		Classic CNN & 1.15M & 81(18) &  64(28) & 65(28) \\
		\hline
		
	\end{tabular}
	\vspace{0.2cm}
	\label{fig:result1}
\end{table}

\begin{figure}[b]
	\centering
	\includegraphics[width=\linewidth]{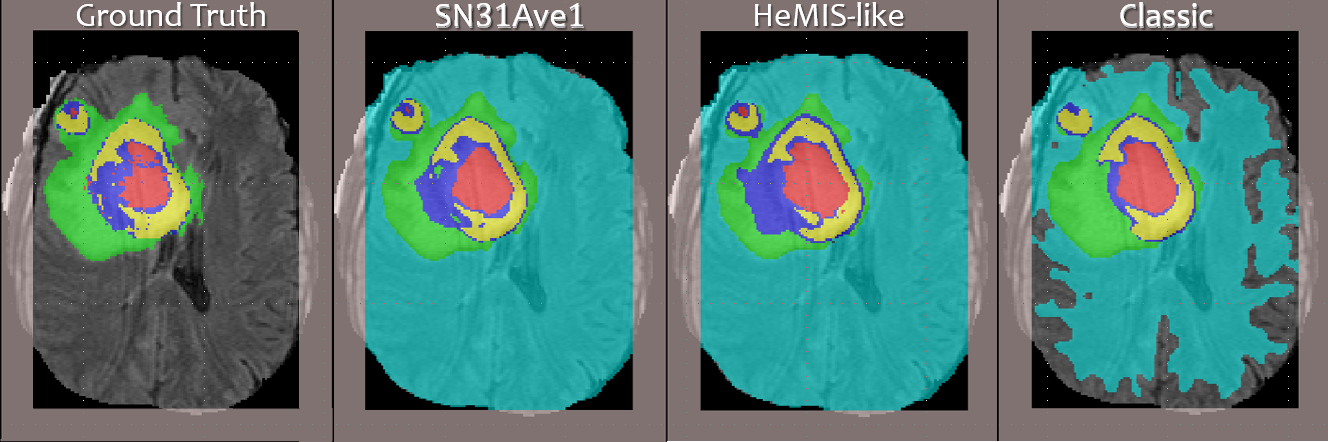}
	\caption{Qualitative comparison of different models output on a particular testing case. Colours correspond to different tissue regions. red: necrotic core, yellow: enhancing tumour, blue: non-enhancing tumour, green: edema, cyan: healthy tissues.}%
	\label{fig:result2}
\end{figure}

\subsubsection{ScaleNet with basic merging and classic CNN.}
We compare three merging strategies (averaging: "SN31Ave2", maxout: "SN31Max2" and concatenation: "Classic CNN"). 
To be as fair as possible, we carefully choose the size of the kernels so that the maximum receptive field remain the same across all architectures.
Quantitative Dice score results Tab.~\ref{fig:result1} show that both SN31Ave2 and SN31Max2 outperform Classic CNN on the segmentation of all tumour region. SN31Ave2 outperforms SN31Max2 for core tumour and get similar results on whole tumour and enhanced tumour.
%

We compare ScaleNets with resp. 1, 2 or 3 scalable multimodal layers before averaging (resp. named "SN31Ave1", "SN31Ave2", "SN31Ave3"). The results reported on Tab.~\ref{fig:result1} show similar performance for all of those models. 
This suggests that a short backend is enough to get a modality-agnostic sufficient representation for Gliomas segmentation using T1, T1c, FLAIR and T2.
Furthermore, SN31Ave1 outperforms Classic CNN on all tumour regions ($p \le 0.001$).

Qualitative results in a testing case with artifact deformation (Fig.~\ref{fig:result2}) and the decreasing of Dice score standard deviation for whole and core tumour (Tab.~\ref{fig:result1}) demonstrate the robustness of ScaleNets compared to classic CNNs and show the regularisation effect of the proposed scalable multimodal layers Fig.~\ref{fig:layers}.

\subsubsection{Comparison to state-of-the-art.}
We validate the usefulness of the cross $M$-feature layers by comparing our proposed network to an implementation of ScaleNets aiming at replicating the characteristics of the HeMIS network \cite{havaei2016hemis} by removing the cross $M$-feature layers.
We refer to this latest network as HeMIS-like. Dice score results in Tab.~\ref{fig:result1} illustrate improved results on the core tumour ($p \le 0.03$) and similar performance on whole and active tumour.
Qualitative comparison in Fig.~\ref{fig:result2} clearly confirmed this trend.

We compare our SN31Ave1 model to the state-of-the-art. The results obtained on Leaderboard and Challenge BraTS'13 dataset are reported in Tab.~\ref{tab:results} and compared to the BraTS'13 Challenge Winners listed in \cite{menze2015multimodal}. We achieved similar results with no need of post-processing.

\begin{table}[t]
	\centering
	\caption{Dice score on Leaderboard and Challenge against BraTS'13 winners.}
	\begin{tabular}{l|ccc|ccc}
		\hline
		 & \multicolumn{3}{c}{\bf Leaderboard} & \multicolumn{3}{c}{\bf Challenge}\\
		\hline
		Method & \multicolumn{1}{c}{~~Whole~~}   & \multicolumn{1}{c}{~~Core~~}   & \multicolumn{1}{c}{~~Enhanced~~}  & \multicolumn{1}{c}{~~Whole~~}   & \multicolumn{1}{c}{~~Core~~}   & \multicolumn{1}{c}{~~Enhanced~~}  \\ 
		\hline
		Tustison & \bf79& \bf 65& 53& 87& \textbf{78}& \textbf{74} \\
		Zaho     & \bf79& 59& 47& 84& 70 & 65 \\
		Meier    & 72&  60& 53& 82& 73 & 69 \\
		SN31Ave1 & 77&  64& \bf56& \bf 88& 77 & 72 \\
		\hline
		
	\end{tabular}
	\vspace{0.2cm}
	\label{tab:results}
\end{table}

%


\section{Conclusions}

We have proposed a scalable deep learning framework that allows building more reusable and efficient deep models when multiple correlated sources are available.
In the case of volumetric multimodal MRI for brain tumour segmentation, we proposed several scalable CNNs that integrate smoothly the complementary information about tumour tissues scattered across the different image modalities.
ScaleNets impose a sparse structure to the backend of the architecture where cross features and cross modalities transformations are separated.
It is worth noticing that ScaleNets are related to the recently proposed implicit Conditional Networks \cite{ioannou2016decision} and Deep Rooted Networks~\cite{ioannou2016deep} that use sparsely connected architecture but do not suggest the transposition of branches and grouped features. Both of these frameworks have been shown to improve the computational efficiency of state-of-the-art CNNs by reducing the number of parameters, the amount of computation and increasing the parallelisation of the convolutions.

Using our proposed scalable layer architecture, we readily adapted a compact network for brain parcellation of monomodal T1 into a multimodal network for brain tumour segmentation with 4 different image modalities as input. Scalable structures, thanks to their sparsity, have a regularisation effect. Comparison of classic and scalable CNNs shows that scalable networks are more robust and use fewer parameters while maintaining similar or better accuracy for medical image segmentation. 
Scalable network structures have the potential to make deep network for medical images more reusable. We believe that scalable networks will play a key enabling role for efficient transfer learning in volumetric MRI analysis.

\subsubsection{Acknowledgements.}
This work was supported by the Wellcome Trust (WT101957, 203145Z/16/Z, HICF-T4-275, WT 97914), EPSRC (NS/A000027/1, EP/H046410/1, EP/J020990/1, EP/K005278, NS/A000050/1), the NIHR BRC UCLH/UCL, a UCL ORS/GRS Scholarship and a hardware donation from NVidia.


\bibliographystyle{splncs03}
\bibliography{JNAbrv,bibliography}

\end{document}